\documentclass[letterpaper, 10 pt, journal]{ieeetran}  

\usepackage[utf8]{inputenc} 
\usepackage[T1]{fontenc}    
\usepackage{url}            
\usepackage{booktabs}       
\usepackage{amsfonts}       
\usepackage{nicefrac}       
\usepackage{microtype}      
\usepackage{color}
\usepackage{subfigure}
\usepackage{caption}
\usepackage{tikz}
\usepackage{paralist}
\usepackage{booktabs} 
\usepackage{placeins}
\usepackage{color}

\usepackage{amsmath}
\usepackage{algorithm}
\usepackage{algpseudocode}
\usepackage{array,booktabs}
\usepackage{flushend}

\usepackage{multirow}

\usepackage{tabularx}
\usepackage{adjustbox}
\usepackage{rotating}

\makeatletter
\let\NAT@parse\undefined
\makeatother
\usepackage{hyperref}
\usepackage{xcolor}
\usepackage{wrapfig}

\title{\LARGE \bf
	Gradient-free Multi-domain Optimization for Autonomous Systems
}

\author{Hongrui Zheng$^1$, Johannes Betz$^1$, Rahul Mangharam$^1$\vspace{-1.5\baselineskip}
		\thanks{$^1$Authors are with the Department of Electrical and Systems Engineering, University of Pennsylvania, Philadelphia, USA}
		\thanks{$^1$\{\texttt{hongruiz, joebetz, rahulm}\}@seas.upenn.edu}}

\begin{document}

	\maketitle
	\thispagestyle{empty}
	\pagestyle{empty}

	\begin{abstract}
Autonomous systems are composed of several subsystems such as mechanical, propulsion, perception, planning and control. These are traditionally designed separately which makes performance optimization of the integrated system a significant challenge.   
	In this paper, we study the problem of using gradient-free optimization methods to jointly optimize the multiple domains of an autonomous system to find the set of optimal architectures for both hardware and software.
	We specifically perform multi-domain, multi-parameter optimization on an autonomous vehicle to find the best decision-making process, motion planning and control algorithms, and the physical parameters for autonomous racing. 
	We detail the multi-domain optimization scheme, benchmark with different core components, and provide insights for generalization to new autonomous systems. In addition, this paper provides a benchmark of the performances of six different gradient-free optimizers in three different operating environments.
    Our approach is validated with a case study where we describe the autonomous vehicle system architecture, optimization methods, and finally, provide an argument on gradient-free optimization being a powerful choice to improve performance of autonomous systems in an integrated manner.

\end{abstract}

	\section{INTRODUCTION}



The design of Cyber-Physical-Systems (CPS) includes a wide variety of engineering areas such as mechanical, electrical, control and decision making.
\textit{Autonomous Systems}, within CPS, has a wide variety of applications: Unmanned aerial vehicle (UAV), unmanned underwater vehicles (UUV), and autonomous vehicles (AV) to name a few.
For an Autonomous System to function properly, core components such as autonomy algorithms (perception, planning, and prediction) \cite{Pendleton2017}, physical components, and low level control algorithms must work together.
However, to find the optimal configuration for a good performance of an Autonomous System, there are multiple challenges that system designer must overcome.
First, the right combination of core hardware components must be determined. This task is hard to solve because for each of these core components, there are usually more than dozens of either continuous or categorical parameters that could change the behavior of each component. This creates a massive combinatorial problem that cannot be solved manually by tuning for different parameters. This problem becomes even harder when evaluating a single combination is computationally, or manually expensive.
In addition, clear measurements that define the overall performance of the autonomous system and in addition display a comparability for other autonomous system developer is needed.
Lastly, due to the cyber-physical nature of Autonomous Systems, any simulation created will have to use models that have a certain discrepancy to real world behavior (e.g. sensor models, vehicle dynamics models), which creates a sim-to-real gap that leads to different performances on real hardware. In addition, choosing an objective function that might differ between simulation and real world might lead to unexpected behavior that ends up exploiting the simulation dynamics. 
In order to overcome these issues we want to show that a deterministic and embarrassingly parallel optimization pipeline can be set up that derives optimal hardware and software configuration for autonomous systems. As an effort to push forward more generalizable optimization methods suitable for a wide range of systems, this work has three primary contributions.
\begin{enumerate}
	\item We provide a scheme for gradient-free, multi-domain optimization of an autonomous system and provide case studies with specified target hardware, modular software, and a calibrated simulator of an autonomous race car.
	\item We provide thorough benchmark evaluating the performance of different optimization techniques with focus on gradient free methods and describe a methodology for tuning a high-dimensional set of hyperparameters across multiple domains.
	\item We provide insights in adapting and optimizing heterogeneous components of new autonomous systems in different domains.
\end{enumerate}

In what follows, we demonstrate that the proposed multi-domain optimization is capable of finding optimal configurations for both hardware and software of an autonomous system.
In Section \ref{sec:method}, we describe the problem statement and the design of our optimization pipeline.
In Section~\ref{sec:related} we provide the state of the art and place our solution in context with previous approaches.
In Section \ref{sec:exp}, we describe the experimental setup and analysis of generated results.
Lastly, in Section \ref{sec:conclusion}, we detail future work and conclusions.

	\section{METHODOLOGY}
	\label{sec:method}

\begin{figure*}[h]
	\centering
	\includegraphics[scale=0.2]{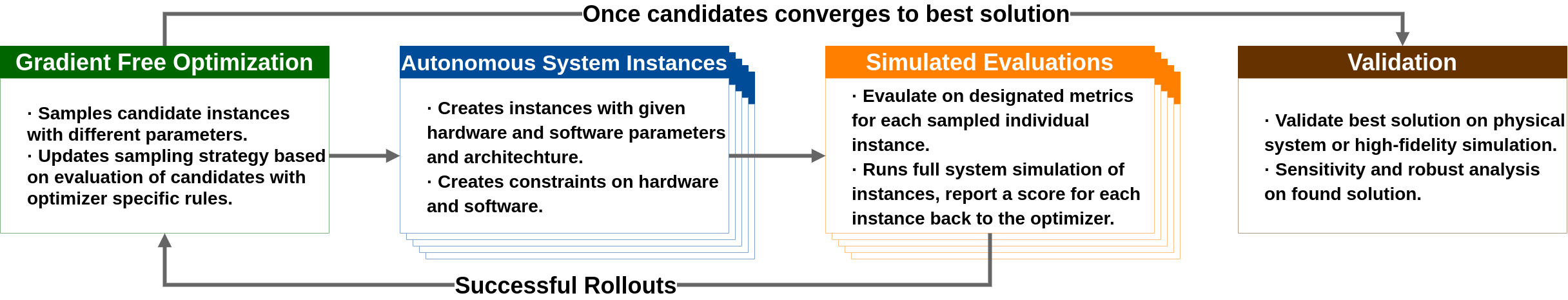}
	\caption{Proposed pipeline for autonomous systems optimization.}
	\label{fig:pipeline}
\end{figure*}

To solve the problem of creating an optimized autonomous system we propose the pipeline displayed in Figure \ref{fig:pipeline}.
In the first step, the head node of the optimization samples candidate instances of solutions with different parameters according to some underlying sampling strategy.
In the second step, a fixed number of parallel workers are created to receive the sampled candidate solutions and create an Autonomous System simulation instance for each of them. So far we have explored AVs, UUVs and UAVs as autonomous systems that are presented as standalone autonomous systems. An extension regarding e.g. connected systems can be done in the problem formulation but not was explored yet.
In the third step each of the Autonomous System instances is now evaluated in a full system simulation. This simulation provides the environment, the virtual sensors, and the Autonomous System dynamics. After simulating each instance with the given candidate, the workers collects metrics on the performance of each instance and reports its findings back to the optimization head node where the iteration starts again for the next generation. This also updates the underlying sampling strategy that the head node uses.
Once all candidates converged to a best solution, the optimization halts and provides a final candidate with the most performant combination of parameters. The solution could then be validated in either a high fidelity simulation environment or with real hardware.

\subsection{Problem Statement}
We first loosely define an autonomous system as a system with:
\begin{inparaenum}
	\item intricate interaction between multiple components that produces complex dynamics;
	\item variable hardware and/or software parameters that influence the overall performance on given metrics.
\end{inparaenum}
Some examples of an autonomous system include: buildings with energy efficiency optimization, self-driving cars, unmanned aerial and underwater vehicles.
The goal of the optimization framework is to find the best combination of parameters in the search space that increase the overall performance of the autonomous system. Mathematically, we define the objective of the optimization problem as:
\begin{equation}
	\begin{array}{ll}
		\operatorname{minimize} & f(\theta)\\
		\operatorname{\text{subject to}} & \theta \in \Theta
	\end{array}
\end{equation}
Where $f(\Theta):\mathbb{R}^n\rightarrow\mathbb{R}$ is the objective function that measures the system's performance on a metric, and a lower function value corresponds to a higher overall system performance. $\Theta$ is the concatenation of all feasible search space of all parameters being tuned. $\theta$ is a candidate sampled from $\Theta$. We do not assume the existence of a gradient and the convexity of the objective function. By defining the problem as stated we can use the proposed approach for optimizing autonomous systems of all kinds.

\subsection{Search Space}
\label{sec:search_space}
We divide the search space generally into three sub-spaces: \textit{Physical} parameters, \textit{High-level decision-making} parameters, and \textit{Low-level control} parameters.

The physical parameters $\Theta_p$ are defined as the parameters that can be physically altered in an autonomous system. For example the mass of the system, the position of the center of gravity (CoG), the hardware components of the system etc. We further provide guidance on how to limit the range of these parameters, and how to initialize these parameters in the search in section \ref{sec:discuss}

The high-level decision-making parameters $\Theta_{dm}$ are defined as the parameters that governs the high level planning decisions of the autonomous system. For example, the importance an autonomous system takes to sacrifice robustness for performance (e.g. the magnitude of a slack variable in a constraint), the parameters in a neural network that controls an autonomous system end-to-end, or the creation of a trajectory (path planning) the system is trying to track.

The low-level control parameters $\Theta_{lc}$ are defined as the parameters that affect the low-level controller of the autonomous system. For example, the $k_{p}$, $k_{i}$ and $k_{d}$ parameter in an PID-controller or the optimization parameters in an MPC-controller.

In section \ref{sec:exp}, we provide a real world case study where all of the parameters in the concatenated search space for the system are clearly defined and discussed.

\subsection{Evaluation Criteria}
The evaluation criteria depends on the specific system that's being optimized. We assume the evaluation criteria to be a function that maps from the parameter search space $\mathbb{R}^n$ to a single or a vector of real numbers. The optimization process doesn't necessarily optimize in only one dimension that measures performance. As long as the objective function satisfies the requirements, there could be multiple aspect of the autonomous system being optimized. For example, we can optimize the performance of an autonomous vehicle in terms of it's fuel efficiency, in terms of power output, but also in terms of the computation efficiency of the on-board software. For a given set of parameters $\theta$ sampled from $\Theta$, we calculate the objective function $f(\theta)$ through simulation of the autonomous system, then the obtained score is returned to the optimizer to push the optimization forward for the next sample.

In section \ref{sec:exp}, we give clear and well-defined evaluation criteria on our real world case study.

\subsection{Optimization}
Since we do not assume the convexity and smoothness, the upper bound of the dimension, and the existence of closed-form expression of the underlying dynamics, we use gradient-free, black-box optimization to solve the problem. A potential pitfall of this method of optimization on our target autonomous system is the potential existence of local optima, which creates inherent tension between exploration of the search space and exploitation of solutions. We discuss and analyze this failure mode in section \ref{sec:discuss} and \ref{sec:exp}.

In our experiments, we use the following optimization algorithms:

\begin{enumerate}
	\item \textbf{CMA} (Covariance Matrix Adaptation) is an evolution strategy \cite{hansen2001completely}. For each generation, CMA samples potential solutions in a multivariate normal distribution defined by a mean and covariance. The mean of the distribution is updated to maximize the likelihood of previously successful solutions based off of a fitness function that evaluates the quality of solution. The algorithm keeps track of two evolution paths (isotropic and anisotropic) based on the previous and current mean. The isotropic evolution path exploits the covariance matrix which helps the algorithm choose a better distribution for the next generation. The anisotropic path is used to control the step size of the algorithm for an optimal convergence. After iterating through a number of generations the algorithm eventually converges on an optimum.
	CMA is an excellent choice when the objective function is non-convex, multi-modal and noisy and when the computational budget is large.


	\item \textbf{TwoPointsDE} and \textbf{NoisyDE} are implementations of the Differential Evolution (DE) \cite{price2013differential} algorithm. DE consists of four phases: initialization, mutation, recombination, and selection. In initialization, a population is sampled uniformly according to the constraints of the parameters. Next, each of the genomes goes through mutation where a donor genome is created by first randomly choosing a number of distinct genomes (not including the current genome going through mutation), adding the weighted differences of a number (depending on implementation) of these chosen genomes to another chosen genome. Then in recombination, a trial specimen is created. At each parameter in a single genome, a value is chosen either from the target genome without mutation, or from the donor genome created in mutation, with some chosen probability. And lastly in selection, the trial genome goes through evaluation. And the target genome for next generation is selected by choosing whichever genome scored higher in the evaluation, the current target genome, or the created trial genome.
	Both TwoPointsDE and NoisyDE are specific implementations of DE where the settings are different in the mutation step. TwoPointsDE refers to using four random distinct genomes for the weighted difference, and the best scoring genome as the base for addition. And NoisyDE refers to using two random distinct genomes for the weighted difference, and the mean of individual genomes that scores better than than median is used as the base for addition.

	\item \textbf{PSO} (Particle Swarm Optimization) \cite{kennedy1995particle, shi1998modified} creates candidate solutions as particles with inertia, and moves the particles around in the search space according to a dynamics equation considering the particle's position and velocity. Each particle is influenced by its local best known position, and is also guided towards the best known positions in the search space. At each generation, the velocity of each particle is created by combining the current velocity of the particle with randomly weighted components calculated using the difference in position between the particle and both the local and global best particle.


	\item \textbf{OnePlusOne} \cite{styner2000parametric} iterates randomly selecting a new position in parameter space, which is controlled by a probability function centered at the current location. The probability function grows or shrinks by un update rule that considers local information.


	\item \textbf{RandomSearch} is the classical random search baseline. All individuals are generated randomly in the search space and used as a comparison for other optimizers.
\end{enumerate}

	\section{RELATED WORK}
	\label{sec:related}
The optimization process proposed in this paper is an effort to generalize and improve the toolchain presented in \cite{OKelly2020}. In its original form, the toolchain is specific to the autonomous race car scenario where the parameter space and application is hardly generalizable. The original optimization core of the toolchain is also a simplified CMA-ES without the option to control explore and exploit. Our work builds upon \textsc{TunerCar} to extend the applicability to all Autonomous Systems. We also extend the options for the optimization core where different algorithms could be utilized to best suit different target systems.

There are several different areas of work that attempts to optimize for the overall performance of a system. We roughly split them into the four following categories.

\subsection{CPS Design Space Exploration}

In \cite{Buini2015} the authors present a design space exploration (DSE) approach for CPSs that emphasizes the variabilities of the physical subsystem and control aspects of the system. As a case study the authors choose an inverted pendulum  with a range of physical and cyber settings.
\cite{Muhleis_2011} considers control performance as design objective and presents a control performance analysis. The approach is based on a co-simulation of high level models of plants and a virtual prototype of the controllers in the system.
In \cite{campos2019task}, an approach to automatically synthesize both the design and control for modular robots is displayed. This synthesis is mainly based on the task the robot should fulfill and the results show that such a synthesis can outperform genetic algorithm optimizations.
In the work of \cite{Vanommeslaeghe2019}, the authors show how domain knowledge can be used to guide the design-space exploration process for an advanced control system and its deployment on embedded hardware. They choose this integrated approach because the parameters in both the control and embedded domain can be chosen, evaluated and optimized to have a performant solution in both domains although the combined design space becomes vast.
\cite{Bradley2015} surveys the run-time cooperative optimization of cyber and physical systems and introduces two methods for balancing cyber and physical resources in a step toward holistic co-design of CPS.
In \cite{Lattmann2012}, a Cyber-Physical Modeling Language (CyPhyML) to support trade studies and integration activities in system-level vehicle designs is presented. This setup is used to test simulation models in terms of generic assemblies over the entire design space. The performance of a design instance once automated design space exploration is complete is evaluated.
Lin et al. \cite{Lin2018}  presents a formalization of the design constraints of building an autonomous driving system in terms of performance, predictability, storage, thermal and power. This work was presented only with focus on the computation hardware only (ECU implemented with CPU, GPU or FPGA).
The same task was displayed in \cite{Collin2019} where computation platforms are optimized under latency and cost.

\subsection{Optimization in System Control}
The final reliability and safety of the autonomous vehicle is mainly connected to a well tuned vehicle controller. Seeking the optimal control parameters for e.g. a PID is a well-known problem. Techniques for solving this issue can be found in \cite{Toscano2005}, \cite{Killingsworth2006}.
Kim et al. \cite{Kim2012} propose to optimize the different parameters of an autonomous car controller by using self-adaptive evolutionary strategies. By using a vehicle simulation environment they conclude that their strategy is competitive to manual controller tuning and design.
Castillo et al. \cite{Castillo2012} present an ant colony optimization and particle swarm optimization to optimize the parameters of a fuzzy logic controller in an autonomous robot.
Similar research was done by Weijia et al. \cite{WeijiaMa2011} to improve the PID controller parameters of a remote controller system. Both publications derive that the optimization techniques lead to better parameters that make the system more robust and overshoot less.
A co-simulation tool-chain with different simulation environments was developed by Dai et al. \cite{Dai2017} to optimizes the control parameters of autonomous vehicle software based on predefined metrics. Although their environment provides extra dimension of fidelity in simulation, e.g. additional sensor simulation, it is not stated how computationally efficient the simulations are to create valuable results. 
With a specific focus on a speed controller only, Naranjo et al. \cite{Naranjo2020} presents an approach using meta-heuristics to optimize the controller parameters. By proposing a global error function the authors use heuristic optimization techniques to find the controller parameters. The results present show controller parameters that have improved accuracy in tracking the speed requirements of the vehicle.
In \cite{Nguyen2008} an approach for local Gaussian process regression is presented to approximate the dynamics of the robots by measuring data and to adapt the control parameters afterwards. These approaches are then applied by different authors like \cite{Wischnewski2020} \cite{Kabzan2019} in the field of path planning and control for autonomous vehicles.

\subsection{Population-based Optimization}
The gradient-free optimization used in our approach could be considered as black-box optimization methods. And most of the algorithms we consider makes use of a generate-update paradigm where a population is generated from a strategy and the strategy is then updated from the performance of the population. We therefore give a short overview of related research in this field.
In \cite{Belter2009}, population-based Methods for the identification and optimization of a robot model are explored.
Similar methods and techniques have recently been utilized as population-based training (PBT) \cite{jaderberg2017population}. Instead of using grid search to train neural networks of a certain architecture with a certain set of hyperparameters from start to end, PBT performs more thorough and faster search by simultaneously exploring multiple designs in combination with hyperparameters at the same time. Less performant variants are terminated early, and descendant variants are mutated from more performant ancestors so that computational resources are not wasted on training flawed network designs.
Similar approaches can be seen at \cite{conti2018improving} and \cite{Shauharda2018}.
In both \cite{ha2018worldmodels}  and \cite{schwarting2021deep}, the synthesis of a controller and planner policies are learned through self-play in imagination with a reinforcement learning approach. Both papers do not address the real-world applicability of the approach, specifically with respect to the lack-of realism in the used simulator and vehicle dynamics. This issue is addressed in \cite{Amini2020} where the control policies for an autonomous vehicle are learned based on a data-driven simulation and training engine by using only sparse rewards.
Cantin \cite{MartinezCantin2019} presents a Bayesian optimization technique that is capable of increasing the convergence speed and reducing the number of samples in standard optimization benchmarks, automatic wing design and machine learning applications.
Song et al. \cite{song2012efficient} displays an initialization approach for Q-learning on mobile robots. By creating a mapping between the known environment and the initial values of the Q-table the robots gets a prior knowledge incorporated in the learning system and therefore a better learning foundation.

\subsection{Superoptimization}
The optimization pipeline we proposed is also similar to the concept of \textit{superoptimization} \cite{schkufza2013stochastic,massalin1987superoptimizer,solar2013program, liang2010learning}. The technique of superoptimization searches automatically for the optimal (most performant) sequence of code in an finite search space. This approach is mainly used in the field of compiler optimization \cite{lattner2002llvm} to improve the performance of the transformed program but has been applied to different domains with certain success \cite{vasilache2018tensor,ragan2013halide}.

	\section{EXPERIMENTS}
	\label{sec:exp}

To evaluate the proposed optimization pipeline for autonomous systems, we experiment with a fully defined autonomous system in the context of autonomous racing.
We choose a single vehicle timed trial scenario where the objective of the system is clearly defined to be lowering the lap time for a specific race track. By choosing this case study we can holistically evaluate the proposed pipeline and compare it to other optimization techniques performing the same task.

\subsection{Experimental Setup}
We use an open source 2D vehicle simulation environment described in \cite{okelly2020f1tenth} as the simulation for the system.
It implements a realistic vehicle dynamics based on the single track dynamics model from \cite{althoff2017commonroad}. The simulation is designed to be deterministic so experiments are reproducible with the same set of inputs. The physics engine is explicitly stepped to enable faster than real time executions. And the design of the simulation is also modular, which enables interchangeability of different race tracks and planning strategies for the autonomous race car. In addition, the simulation environment also provides a LiDAR simulation and therefore enables experiments with perception algorithms. 


Next, we define the search space and the evaluation criteria of the problem. In this experimental setup we focus on the optimization of path planning, path tracking as well as the hardware definition of the vehicle.
We follow the definitions in Section \ref{sec:method} and specify each of the definitions. First, we describe the search space of the problem:
\begin{itemize}
	\item \textit{Physical} parameters $\Theta_p$: we alter the mass and the location of the center of gravity of the vehicle.
	\item \textit{High-level decision-making} parameters $\Theta_{dm}$: we alter the trajectory the vehicle follows along the full race track, the minimum and maximum longitudinal velocities of the vehicle on the race track.
	\item \textit{Low-level control} parameters $\Theta_{lc}$: we alter the specific controller parameters (e.g. gain for cross track error control) for the chosen controllers used in the experiment.
\end{itemize}
An example search space is described in Table \ref{tab:space}.

\vspace{-1pt}
\begin{table}[h]
	\centering
	\begin{tabular}{|c|c|c|}
		\hline
		\textbf{Parameter}& \textbf{Type} & \textbf{Size}\\\hline
		\multicolumn{3}{|c|}{\textit{Physical} $\Theta_p$}\\\hline
		Mass (kg) & Float & 1\\\hline
		CoG to Front (m) & Float & 1\\\hline
		\multicolumn{3}{|c|}{\textit{High-level Decision-making} $\Theta_{dm}$}\\\hline
		Lowest Velocity (m/s) & Float & 1\\\hline
		Highest Velocity (m/s) & Float & 1\\\hline
		Waypoint perturb & Float Vector & 100\\\hline
		\multicolumn{3}{|c|}{\textit{Low-level Control} $\Theta_{lc}$}\\\hline
		LQR tracking parameters & Float Vector & 5\\\hline
		
	\end{tabular}
	\caption{Search space for default task of using LQR on Spielberg}
	\label{tab:space}
\end{table}

The evaluation criteria $f$ for this problem is straight forward: We want to minimize the time it took one autonomous racecar candidate to travel twice around the race track.

We use Ray \cite{moritz2018ray} to orchestrate the Map-Reduce \cite{dean2008mapreduce} pattern used to run simulations in parallel. Each Ray remote worker has its own simulation instance, listens to work distributed by the head node and report scores back to the head node. We use Nevergrad \cite{nevergrad} as the optimizer library.

\begin{figure}[h!]
	\centering
	\includegraphics[scale=0.2]{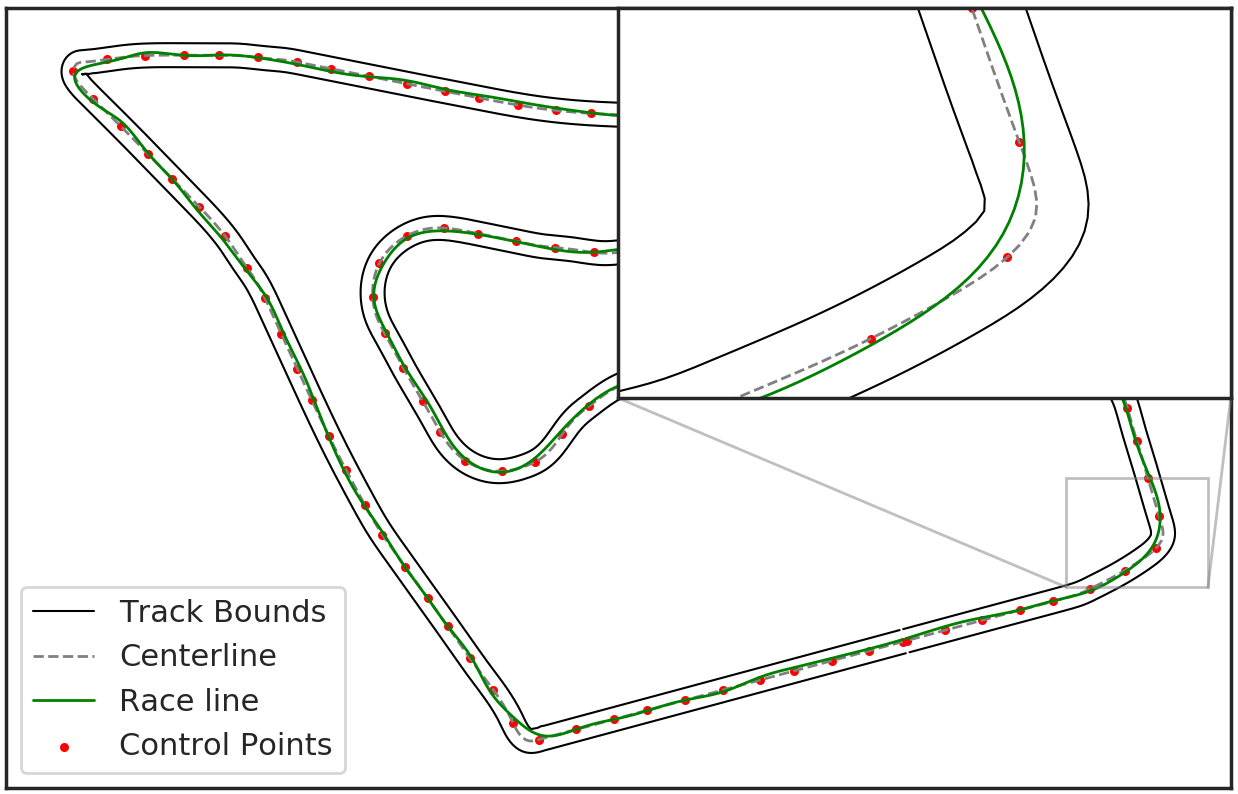}
	\caption{Race line compared to original center line of the track and control points}
	\label{fig:pert}
\end{figure}

When each worker receives a candidate solution to be evaluated, it first update the parameters used in the physics simulation using the sampled mass and location of center of gravity. Next, a controller with the sampled type and parameters are instantiated. Lastly, the worker shifts the control points along a track laterally with the given candidate vector. Finally, race line is created by interpolating a spline through these control points. An example showing a candidate solution of race line created by interpolating the perturbed control points is shown in Figure \ref{fig:pert}.

With this experimental setup, we design the following experiments.
In Section \ref{sec:optimcompare} we compare the convergence performance, stability of optimization and computation budget requirements for different optimizers.
In Section \ref{sec:hypercompare} we compare the effect of choosing different hyper-parameters on the performance of different optimizers.
In Section \ref{sec:generalize} we show the optimization framework's performance in different environments and when different components are used in the autonomous system.
In Section \ref{sec:initcondition} we show that it is crucial to use domain knowledge when creating initial conditions for the optimization search and it is beneficial to set up interpolation strategies to cut down the computation time in the search.
In Section \ref{sec:sensitivity} we perform sensitivity analysis around the solution found by the optimization.
Lastly, in Section \ref{sec:comparemincurv} we compare the solutions found by our method with another optimization method that aims to lower a race car's lap time on track. All optimized parameters for each experiment can be found in Tables \ref{tab:tracker_results} and \ref{tab:all_param}.

\subsection{Comparison of optimizer}
\label{sec:optimcompare}
The objective of this experiment is to compare the convergence performance and computational budget requirement for different optimizers. The race track, controller type for the vehicle and total computation budget are kept the same, with the optimizer used as the only difference between runs. We compare the performance of the following optimizers:
\begin{inparaenum}
	\item CMA
	\item Differential Evolution (DE) with two points cross over
	\item Differential Evolution with noisy recommendation
	\item Particle Swarm Optimization
	\item 1+1 Evolutionary algorithm
	\item Random Search.
\end{inparaenum}
We use a total of 9600 simulations for each run as the computation budget. We also use the default race track \textit{Spielberg} and the default low level controller \textit{Pure Pursuit} \cite{coulter1992implementation}.


\begin{figure}[h]
  \centering
  \includegraphics[width=8.5cm,height=10.5cm,keepaspectratio]{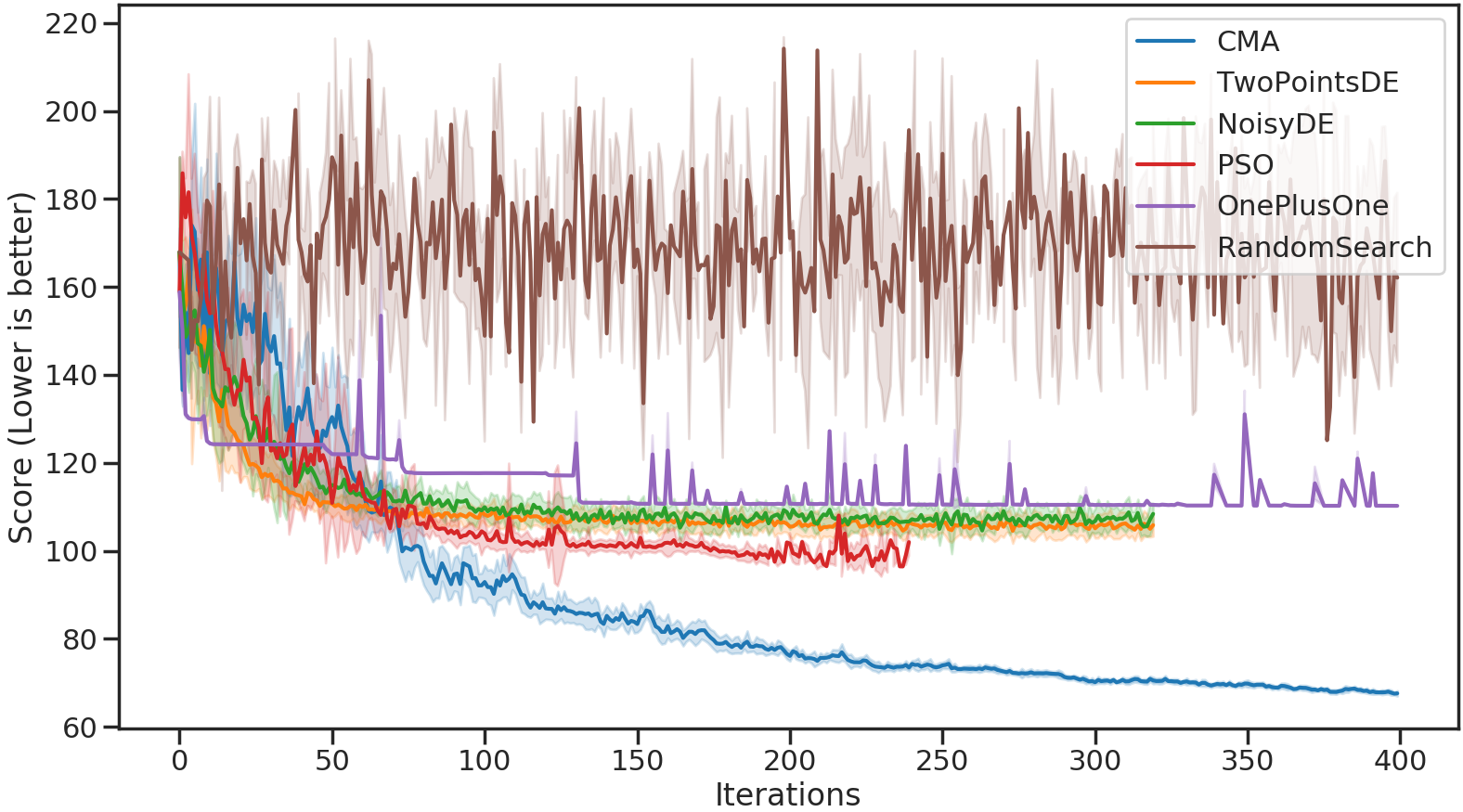}
  \caption{Convergence performance and computational budget requirement of different optimizers}
  \label{fig:exp1_optimizers}
\end{figure}

We show the results of this experiment in Figure \ref{fig:exp1_optimizers} as history of average lap times and standard deviation in each iteration\footnote{Different optimizers have different number of iterations because the number of simulations in each iteration are different. The experiment is controlled to have the same total number of simulations.}. The \textit{Random Search} optimizer performs random sampling in the search space without any search direction. We use this optimizer as the baseline for comparison. As shown in the figure, all other candidate optimizers except for the \textit{Random Search} baseline shows convergent behavior and clear minimization of the objective value. Specifically, we see most of the optimizers converge to the 95 to 115 seconds range and CMA performs the best with the lowest objective value found at 66.69 seconds for 2 laps.


\begin{figure}[h]
	\centering
	\includegraphics[width=8.5cm,height=10.5cm,keepaspectratio]{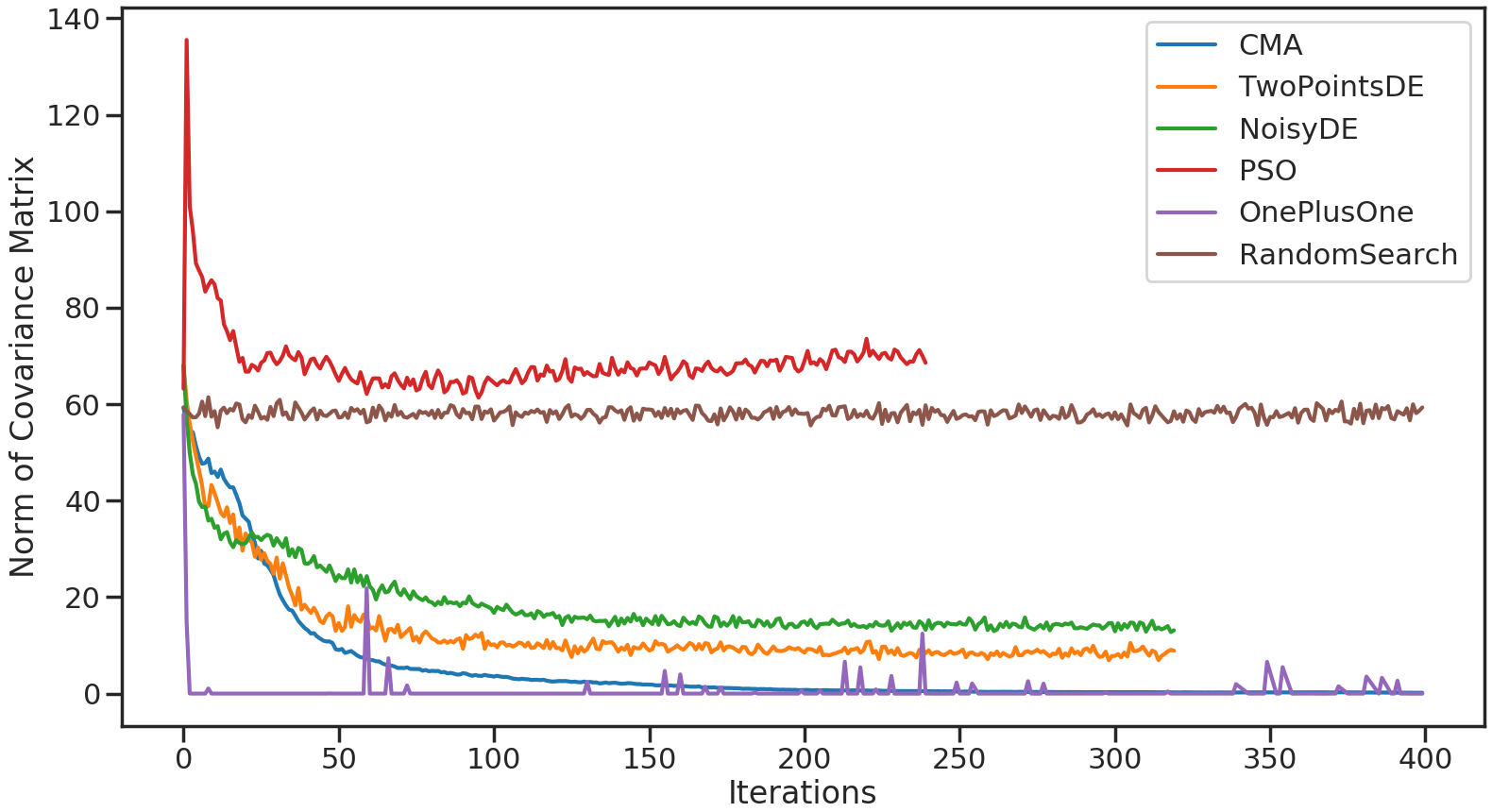}
	\caption{Progress of norm of covariance matrix of different optimizers}
	\label{fig:exp1_normcovmat}
\end{figure}
We also show the convergence of different algorithms as progress of the norm of covariance matrices of the batch of candidates in each iteration for different optimizers. As a baseline, we again see no convergent behavior for Random Search, since the norm of the covariance matrix remains mostly constant throughout the all iterations. For CMA and (1+1), the difference between candidates in a generation vanishes and all individuals converge to a single point. For both optimizers using DE, we see that the population has converged, but not to a single candidate, which is the reason we see non zero norms. And for PSO, the candidates stayed very diverse, even more spread out than the Random Search generations.

We infer that the good performance of CMA on this problem is due to the exploit-explore nature of the algorithm. For CMA, the populations are not confined within the vicinity of the original sampled population. For example, DE recombines genomes in the previous population, which means this operation in nature is an averaging between candidate instances in the optimization. While CMA changes the sampling strategy by modeling the spread of the candidates with a Gaussian distribution. With a larger covariance, the distribution is able to explore outside the range defined by the original sampled candidates. This is also why we see an increase in the covariance between candidates before we see a decrease.


\subsection{Comparison of hyperparameters}
\label{sec:hypercompare}

\begin{figure}[h!]
	\centering
	\includegraphics[scale=0.15]{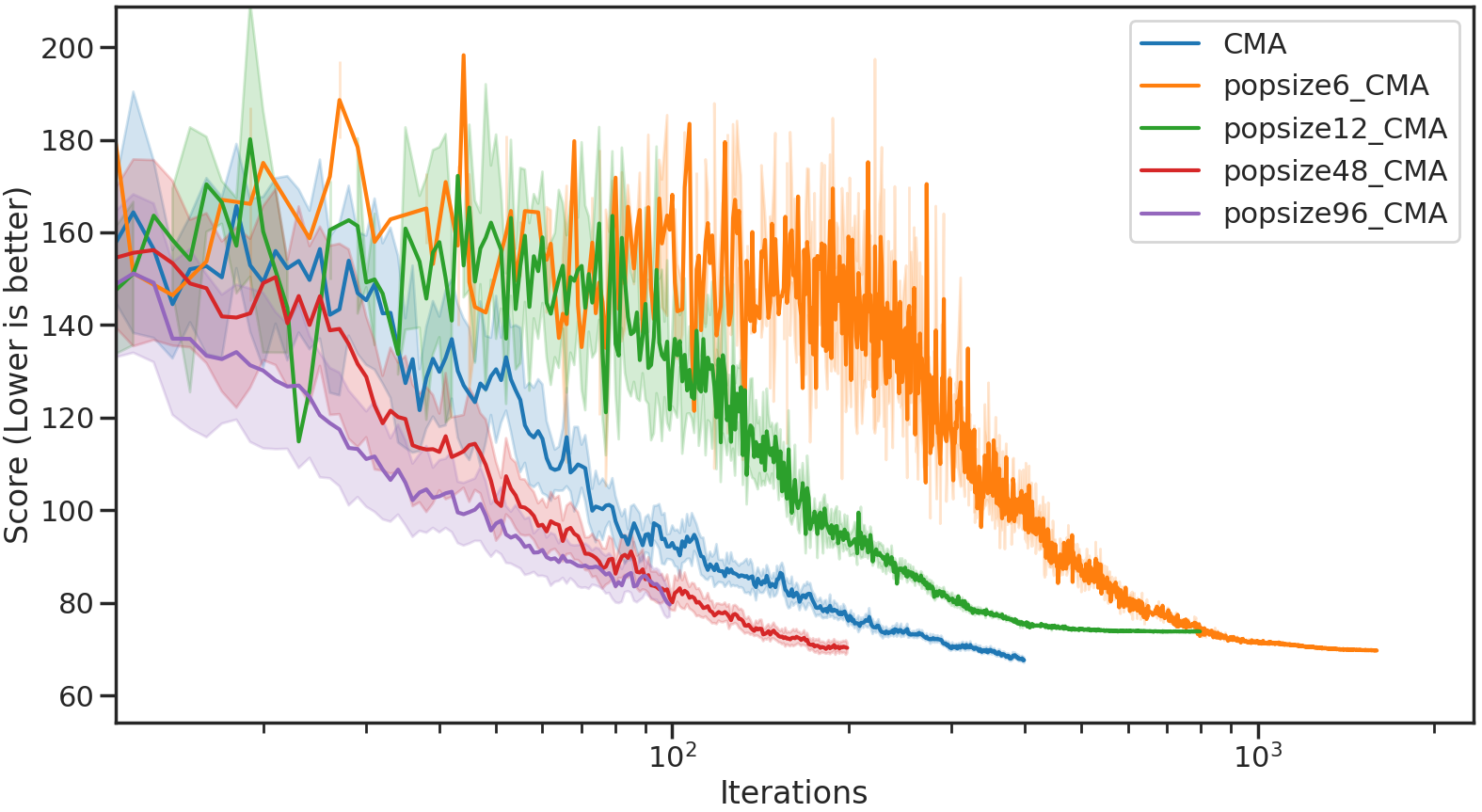}
	\caption{Progression of lap time with differnt population size used for CMA}
	\label{fig:exp2_popsize_CMA}
\end{figure}

Next, we explore different settings used by optimizers and study the effect of changing parameters in certain optimizers.
We first experiment with different population sizes used for different optimizers, while keeping the total computation budget the same.
CMA uses a default population size of 24 in our experiment to match the number of workers used in parallel simulation. We also experiment with population sizes 6, 12, 48 and 96. We show the average lap time and standard deviation in each generation in Figure \ref{fig:exp2_popsize_CMA} and the best lap time after all computation budget is exhausted in Table \ref{tab:cmapopbest}.

\begin{table}[h]
\centering
\begin{tabular}{|c|c|}
	\hline
	\textbf{Population Size} & \textbf{Best Lap time}\\\hline
	6 & 69.65\\\hline
	12 & 73.71\\\hline
	24 & 66.69\\\hline
	48 & 67.9\\\hline
	96 & 73.78\\\hline
\end{tabular}
\caption{Best lap time for CMA using different population sizes.}
\label{tab:cmapopbest}
\end{table}

We see the best lap time was achieved using the default population size of 24, with other settings also finding competitive solutions but worse than using the hyperparameter determined by the default heuristics. We also performed the sane analysis on the two DE based optimization and PSO. The results we see are similar and we show them in Figures \ref{fig:exp3_popsize_2pointsDE}, \ref{fig:exp4_popsize_NoisyDE}, and \ref{fig:exp5_popsize_PSO} in the Appendix.

\subsection{Generalization of Environment and Autonomy Component}
\label{sec:generalize}
\begin{figure}[h]
	\centering
	\includegraphics[scale=0.15]{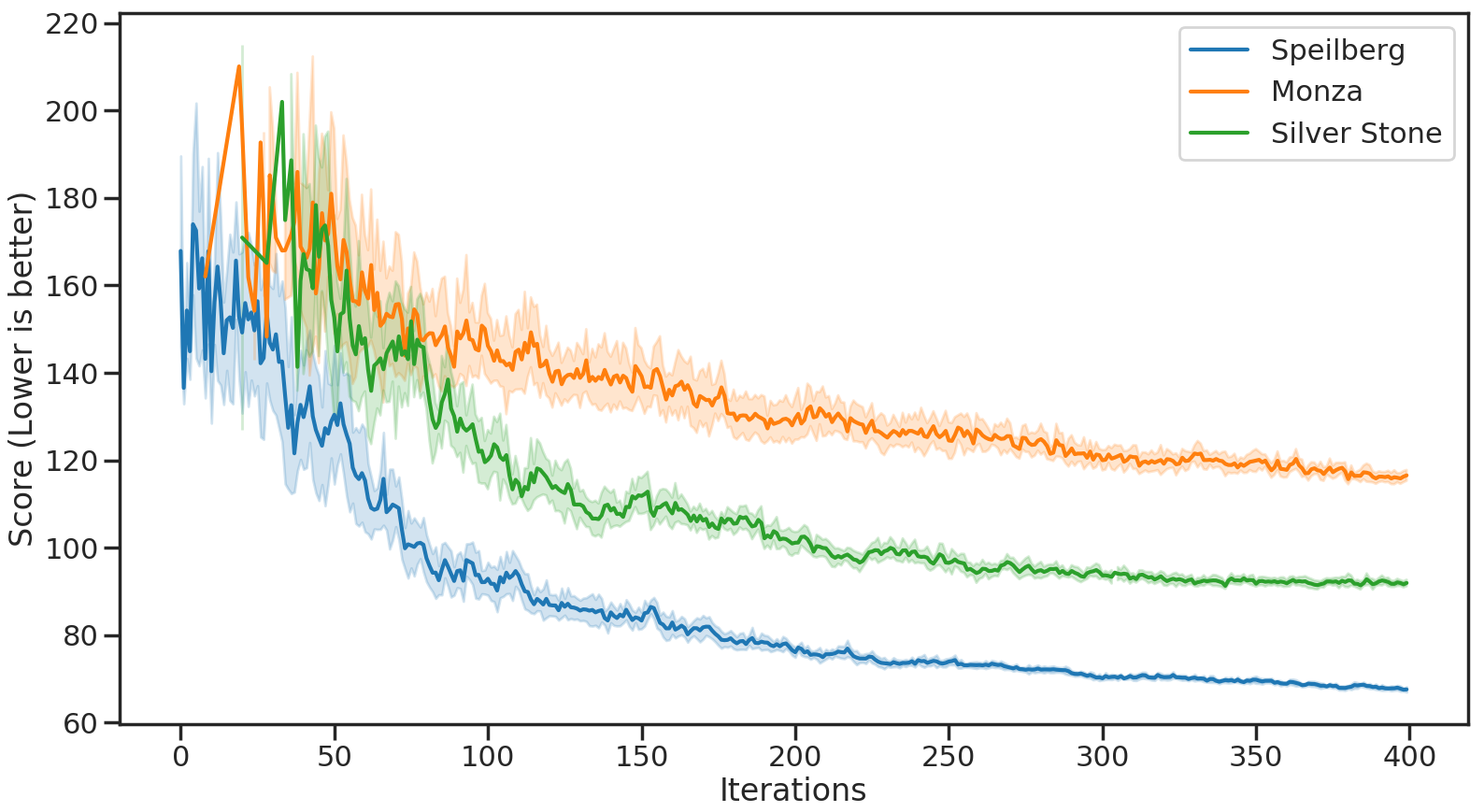}
	\caption{Progression of lap times with CMA on different race tracks.}
	\label{fig:map_compare}
\end{figure}

To display that our proposed framework is able to generalize to different systems and different operating environments, we perform the same optimization using the default setting of CMA on two more different race tracks and with different control strategies.
All three race tracks have unique track layouts displayed in Figure \ref{fig:racetracks} in the Appendix.
The default race track Spielberg is a combination track composed of equal parts straights, sweeper curves and hairpins. While Silverstone is a more curve-heavy race track featuring sweep curves, hairpins and chicanes. Finally, Monza is a high speed race track with mostly straights and three chicanes. The resulting best lap time for each map is shown in Table \ref{tab:cmamaps}, and the progression of lap time for each map is shown in Figure \ref{fig:map_compare}. It is clear that the convergent behavior of the optimization shows across different system operating environments.
\begin{table}[h]
	\centering
	\begin{tabular}{|c|c|}
		\hline
		\textbf{Race Track} & \textbf{Best Lap time}\\\hline
		Spielberg & 66.69\\\hline
		Silver Stone & 89.56\\\hline
		Monza & 113.04\\\hline
	\end{tabular}
	\caption{Best lap time for CMA with Pure Pursuit on different race tracks.}
	\label{tab:cmamaps}
\end{table}

\begin{figure}[h]
  \centering
  \includegraphics[scale=0.15]{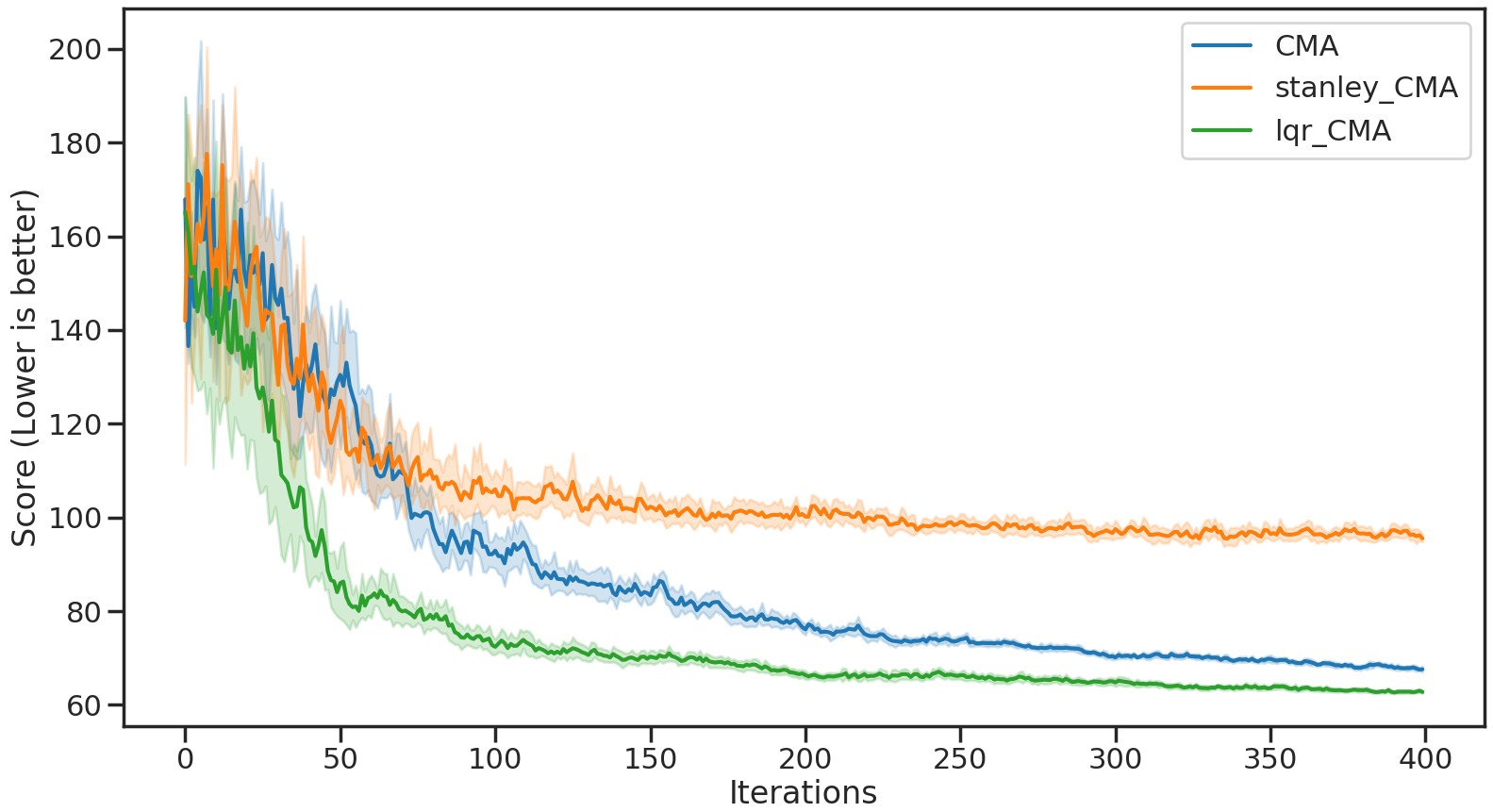}
  \caption{Progression of lap times with different path tracking controller}
  \label{fig:exp8_controller_compare}
\end{figure}

We also experiment with using different autonomy components that fundamentally changes the behavior of the system. We drive the car with three different path tracking controllers: Pure Pursuit \cite{coulter1992implementation}, the Stanley controller \cite{thrun2006stanley}, and a Linear Quadratic Regulator (LQR). Both Pure Pursuit and the Stanley controller are geometric path trackers whereas LQR is an iterative optimization based method. Therefore all three controllers have different sets of parameters to be optimized. We showed the result of the three different controllers successfully being optimized in Figure \ref{fig:exp8_controller_compare}. The best solution found by the different controllers are shown in Table \ref{tab:controller_laptimes}.

\begin{table}[h]
	\centering
	\begin{tabular}{|c|c|}
		\hline
		\textbf{Controller} & \textbf{Best Lap time}\\\hline
		Pure Pursuit & 66.69\\\hline
		Stanley & 92.13\\\hline
		LQR & 62.18\\\hline
	\end{tabular}
	\caption{Best lap time with different tuned controllers on Spielberg.}
	\label{tab:controller_laptimes}
\end{table}

We see that LQR is able to achieve the best result, improving the solution found by using Pure Pursuit by almost five seconds, while Stanley struggled to find a fast solution, but still successfully converged to a solution. This also shows that the pipeline generalizes to systems with different components. The optimized parameters for the three different controllers are shown in Table \ref{tab:tracker_results} in the Appendix.


\subsection{Interpolation the search space and changing initial conditions}
\label{sec:initcondition}
We experiment with changing the initial condition for where the search starts and the constraints on parameters to show the importance of human domain knowledge in bounding the search so that it finds a valid solution easier. We experiment with the default setting using the Spielberg map and the Pure Pursuit controller, while relaxing some of the constrains we impose on the bounds of the variable parameters. We showed the different bounds in Table \ref{tab:search_bounds}. With these few constraint relaxation, the optimizer was not able to find a single solution that's able to complete two laps on track without collision with the same computation budget of 9600 simulations.

\begin{table}[h]
	\centering
	\begin{tabular}{|c|c|c|c|c|}
		\hline
		\textbf{Parameter} & \multicolumn{2}{c|}{\textbf{Original}} & \multicolumn{2}{c|}{\textbf{Relaxed}}\\\hline
		& \textit{Lower} & \textit{Upper} & \textit{Lower} & \textit{Upper}\\\hline
		Mass (kg) & 3.0 & 4.0 & 1.0 & 10.0 \\\hline
		CoG to Front (m) & 0.147 & 0.170 & 0.001 & 0.3 \\\hline
		Lowest Velocity (m/s) & 0.5 & 2.0 & 0.5 & 10.0 \\\hline
		Highest Velocity (m/s) & 6.0 & 15.0 & 10.0 & 20.0 \\\hline
		Lookahead Distance (m) & 0.2 & 2.0 & 0.2 & 10.0 \\\hline
	\end{tabular}
	\caption{Bounds for parameters on original and relaxed constraints.}
	\label{tab:search_bounds}
\end{table}

\subsection{Sensitivity around found solutions}
\label{sec:sensitivity}
\begin{figure}[h!]
	\centering
	\includegraphics[scale=0.15]{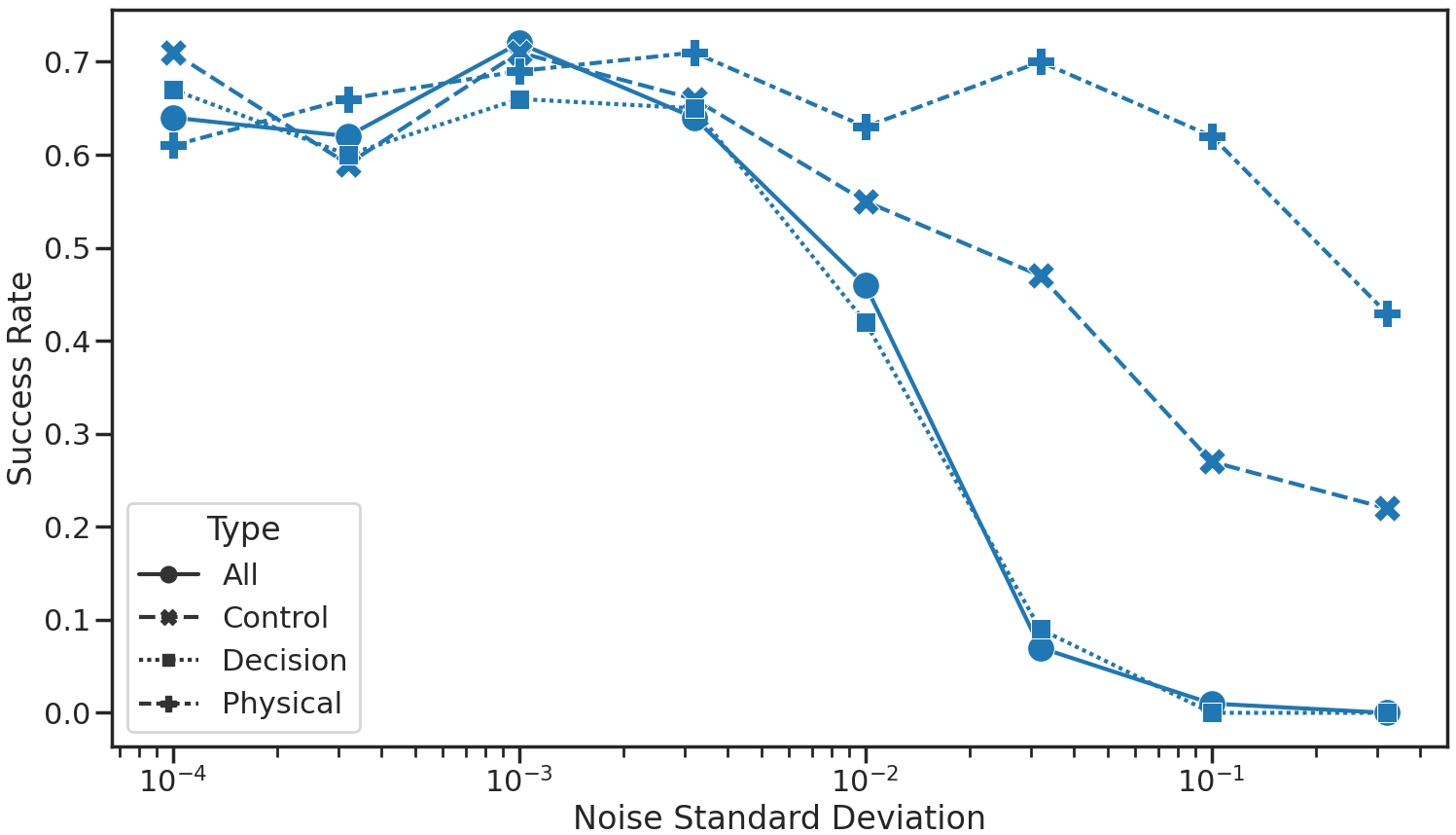}
	\caption{Success rate of solutions v.s. standard deviation of added Gaussian noise}
	\label{fig:sensitivity}
\end{figure}

We examine the points around the found solution of CMA under the default setting. We add Gaussian noises with different standard deviation to the best found solution and run 100 experiments for each scale to find the success rate of shifted solutions being able to complete two laps. We separate the experiment into four sections where noise is added to only a portion of the found solution vector. Figure \ref{fig:sensitivity} shows the results when noise is added to all elements, only the control related elements, only the high level decision related elements, and the physical parameter related elements of the solution vector. As we can see in the figure, the shifted solutions for all experiments maintains between 60\% to 70\% success rate when the noise scale is between $10\text{e}^{-4}$ and $10\text{e}^{-2.5}$. Then when the scale increases beyond $10\text{e}^{-2.5}$, the success rate falls off rapidly and hits zero at $10\text{e}^{-1}$ for \textit{Decision} and \textit{All}. Meanwhile, there are falloffs for both \textit{Control} and \textit{Physical} but the descent is not as steep. We then conclude that the sensitivity of the found solution is highest when the parameters are related to high level decision making, and not as high when it comes to physical and control parameters.

\subsection{Comparison to other methods}
\label{sec:comparemincurv}
\begin{figure}[h!]
	\centering
	\includegraphics[scale=0.13]{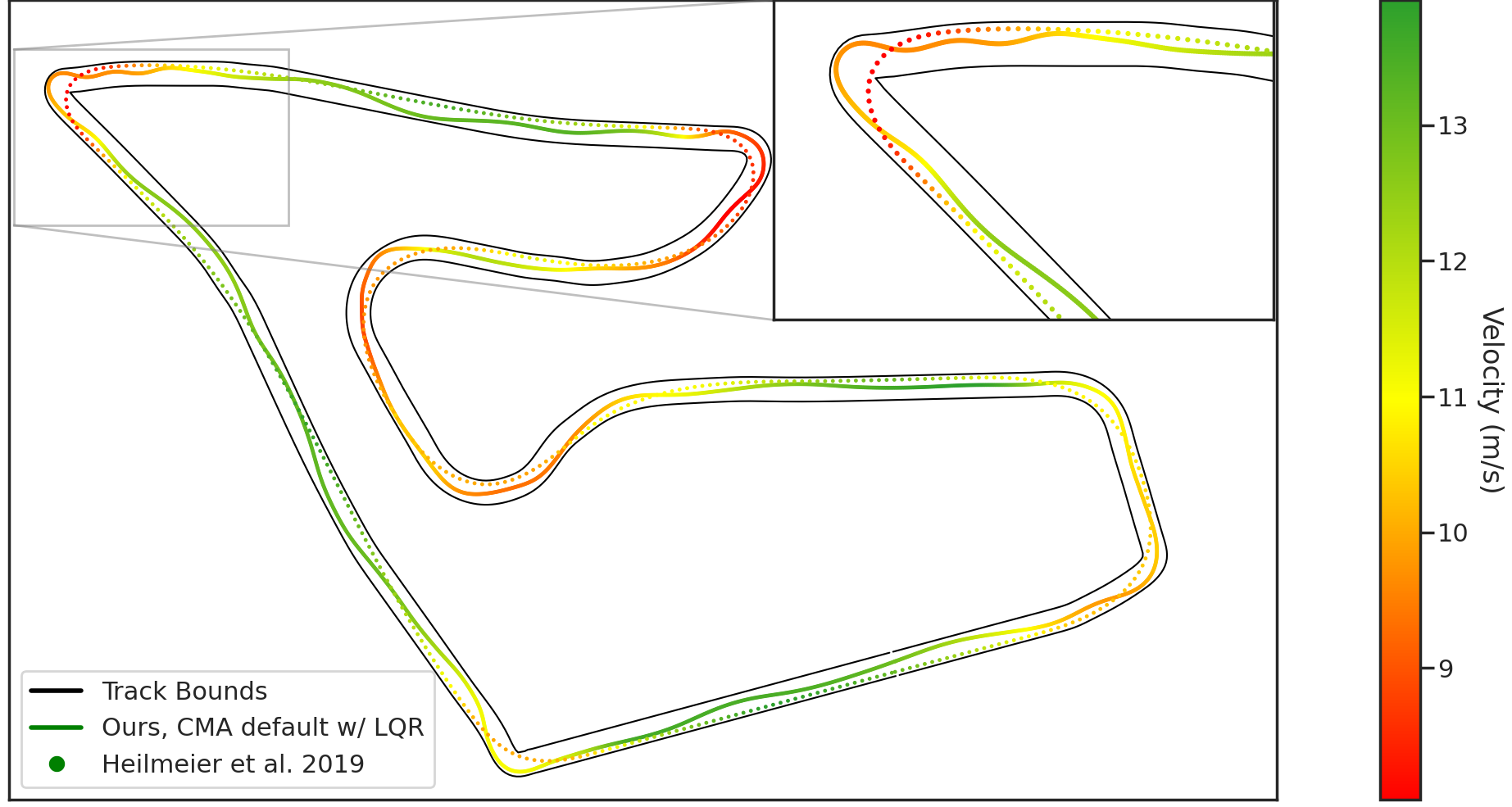}
	\caption{Trajectory comparison between our method and Heilmeier et al. where the color represents velocity and the size of the scattered dot represents the slip angle.}
	\label{fig:comparison}
\end{figure}

In this section, we closely examine the solution found by the proposed optimization framework in this paper and compare it against the solution found on the same track by a more conventional method described in Heilmeier et al.\cite{heilmeier2019minimum}. In this method, a minimum curvature path with the velocity along the path is generated by solving a QP that takes the lateral and longitudinal acceleration limits of the vehicle into account. We adopt this method to our scaled vehicle on the \textit{Spielberg} track to compare with the solution found by CMA under the default setting using the LQR tracking controller. We addittionally compare the trajectory generated by tracking the path and velocity generated by Heilmeier et al. \cite{heilmeier2019minimum} using the LQR tracking controller tuned by our method. \\
With the same maximum allowed velocity on track, same physical vehicle parameter used, same track condition and same number of laps, our method was able to achieve 62.93 seconds for 2 complete laps. Meanwhile, the theoretical lap time from Heilmeier et al. \cite{heilmeier2019minimum} is 67.80 seconds. Using this raceline in the simulation environment with the tuned LQR controller, it achieves 69.34 seconds. Since the LQR controller cannot guarantee navigation without understeer or oversteer, the actual tracked trajectory is slower than the theoretical, minimal curvature lap time. Figure \ref{fig:comparison} shows the comparison between the trajectory produced by the two methods.

\begin{figure}[h!]
	\centering
	\includegraphics[scale=0.25]{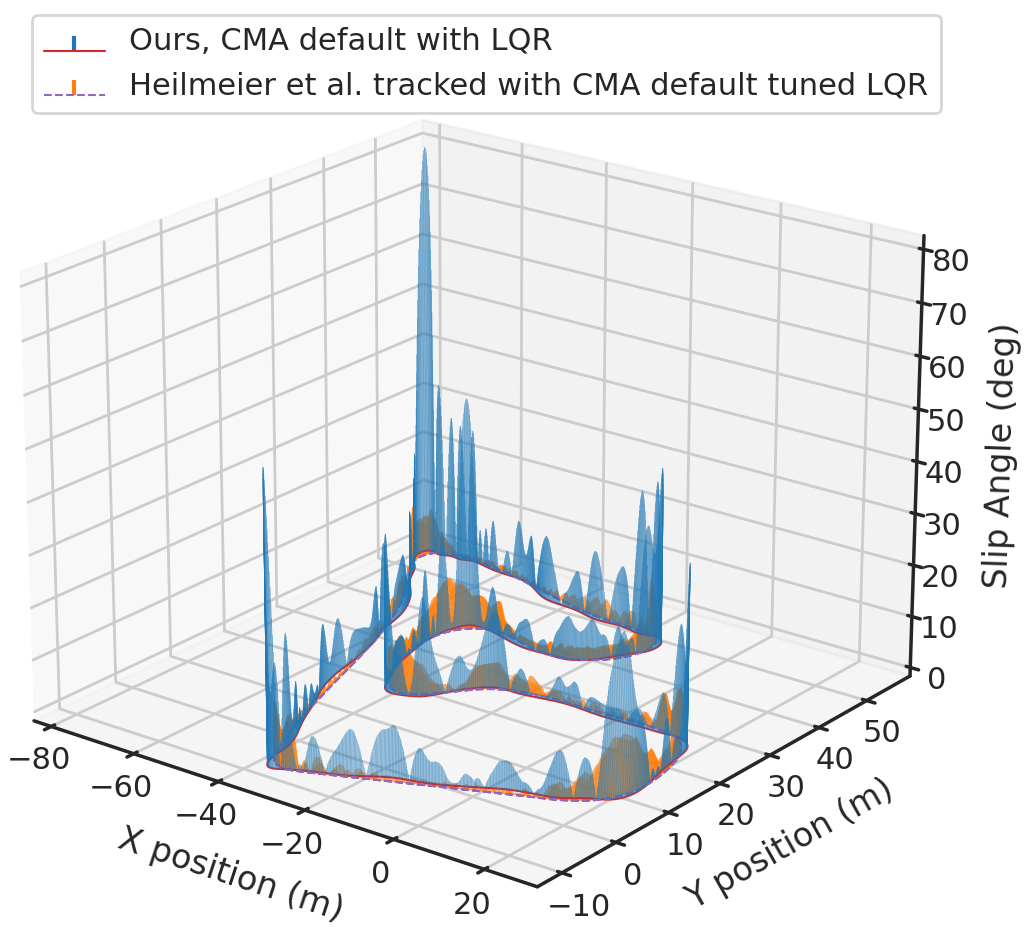}
	\caption{Trajectory of the solution found by using the LQR tracker with the default CMA optimization compared to the trajectory tracked with the same tuned LQR tracker with race line generated by Heilmeier et al.}
	\label{fig:slip_compare}
\end{figure}

A notable difference between the two trajectories is that our solution does not follow the traditional conception of a fast race line. For example, as shown in the figure, our solution does not try to cut the sharp corners by using an inside line and instead drifts around the tight corner while maintaining a velocity about 2 to 3 m/s faster than a traditional race line at the corner. This results in a faster lap overall, but less room for error. Figure \ref{fig:slip_compare} displays an additional comparison of vehicle slip angle between our solution and the trajectory tracked using Heilmeier et al.'s race line and the tuned LQR.

An observation that we can make from this comparison and together with the sensitivity analysis in Section \ref{sec:sensitivity}, we can conclude that our method is able to find a solution that exploits the dynamics of the simulation by finding extremely rare combinations of parameters that is able to navigate around the track with a very low lap time. At the same time, these combinations are extremely sensitive to perturbations and the slightest random noise added to the system results in crashes almost 40\% of the time.

\begin{table}[h]
\centering
\begin{tabular}{|c|c|c|}
	\hline
	\textbf{Method} & \textbf{Path Controller} & \textbf{Best Lap time}\\\hline
	Our approach (CMA) & LQR & 62.93 \\\hline
	Heilmeier et al \cite{heilmeier2019minimum} theoretical & - & 67.80 \\\hline
	Heilmeier et al \cite{heilmeier2019minimum} practical &  untuned LQR & 73.64 \\\hline
	Heilmeier et al \cite{heilmeier2019minimum} practical &  tuned LQR & 69.34 \\\hline
\end{tabular}
\caption{Comparison of raceline methods on Spielberg map}
\label{tab:method_comparison}
\end{table}

	\section{DISCUSSION}
	\label{sec:discuss}
In the previous sections we provided a holistic overview of the proposed optimization framework with a rich set of experiments in the use case of autonomous racing. The results show that our optimization pipeline can indeed optimize an autonomous systems as a whole without explicitly optimizing for specific aspects of the system separately.
We also showed that the pipeline generalizes to different system components (i.e. path tracking algorithms) and different system operating environments (i.e. race tracks).
We observed that the proposed method is able to find solutions that achieves an improved result by exploring solutions that would not be considered by other optimization methods. It is also shown that these solutions are extremely good at the task, while being sensitive to perturbations.
The proposed optimization pipeline for autonomous systems in this paper can be applied to different kinds of autonomous systems and therefore we want to provide additional guidance on how to apply our framework in the development phase of an autonomous system, focusing on
\begin{inparaenum}
	\item how to identify the proper search space for parameters;
	\item techniques that accelerates and bounds the optimization;
	\item selecting the correct optimizer for the problem at hand;
	\item and finally, setting up the infrastructure that takes full advantage of parallelism;
\end{inparaenum}

In addition we will discuss the advantages and disadvantages of our method, along with further improvements that could be made.

\subsection{Identifying the search space}
In Section \ref{sec:search_space}, we described the three sub-spaces that define the parameter set for our given problem. When identifying the search space of a new autonomous system, one should group all variable parameters into these three categories.
After identifying the parameters, one should also identify the constraints on these parameters. This is mainly done by using experts in these fields that have worked in this area before. For example by developing an autonomous vehicle a vehicle width of 5m may be good for the optimization to solve design issues but is impractical considering the width constraint of street lanes.

\subsection{Techniques to accelerate and bound the search}
\begin{enumerate}
	\item It might be counter-intuitive to tighten the constraints in an optimization problem so that a solution can be found easier. Since gradient free optimizations are sampling based, tightening the constraints actually shrinks the search space for the algorithm and a solution within the range will be found faster.
	As shown in Section \ref{sec:initcondition}, having a tighter restriction on the range of values for parameters helps the optimization find valid solutions.
	Again, the usage of human experts in these fields are encouraged to chose the bounds of the values based on their knowledge of the system. For example, the mass and the distance from center of gravity to front axle are both based on realistic values that's possible to achieve on the physical vehicle.

	\item In addition it is important to select parameters for the optimization that have an impact on the performance of the system. For example the mass of the system is more important to the overall acceleration of the vehicle than its CoG. By having a rough idea of the sensitivity of the parameters on the whole autonomous system, it's easier to identify the more impactful parameters and possibly easier to speed up the optimization.
\end{enumerate}

\subsection{Selecting the right optimizer}
The selection of the right optimizer that's suitable for the problem is critical to the success of the optimization. For example, in our autonomous racing use case, the search space for parameters is large, the rollouts of each evaluation is fast, and easily parallelized, thus CMA is an excellent choice for these type of problems since it's explore-exploit scheme is able to explore a wide range of solutions in the search space with a high number of evaluations.

If the simulation of an autonomous system is slow and inefficient, an efficient search algorithm that is able to yield improved solutions with a low number of evaluations such as \textit{OnePlusOne} should be chosen as the optimizer.

Lastly, if multiple solutions that are unique is required, then such as \textit{PSO} should be chosen. Although the solutions did not converge to the best possible, it was able to find improved combination of parameters that are still diverse in the given computation budget.

\subsection{Software infrastructure to support the optimization}
Our optimization scheme takes advantage of parallelism due to the batched nature of evaluations required by the optimizers. A MapReduce \cite{dean2008mapreduce} pattern in the software architecture is recommended for efficient evaluations. The proposed software architecture will be made available online.\footnote{\hyperlink{https://github.com/hzheng40/tunercar}{https://github.com/hzheng40/tunercar}}

\subsection{Advantages and Disadvantages of the proposed optimization pipeline}
The advantage of the proposed pipeline is the holistic view on optimization of an autonomous system. By not restricting what explicit parameters are allowed in the formulation, both hardware and software components can be optimized at the same time. The optimization space as well as the objective function is defined by a handful of parameters and can therefore be applied to different use cases. By using lightweight simulator environments and low overhead for parallelization and optimization, we achieve efficient computation times on consumer computation hardware. For example, with 9600 evaluations, the CMA experiment with Pure Pursuit on Spielberg takes around 25 minutes on a Ryzen 5900X with no GPU requirements. The usage of HPC hardware where more parallel workers can be spawned would speed up the optimization but is not necessary. The pipeline also provides flexibility in terms of what the desired solutions have characteristically. Core components can also be changed to accommodate different system characteristics.

However, we also address the shortcomings of the proposed solution. The current simulation environment we experimented with is mainly constraint for single system scenarios. An interaction with other agents or systems is not provided or planned. 
The approach also did not consider the discrepancy between simulation and the real system. For example, by only defining a single objective value (in our case study: lap time), a fast lap time was achieved but with a very aggressive and inoperable vehicle behavior (high side-slip angle) which would lead to an unstable vehicle behavior on the real system.
Multi-objective optimization is theoretically possible with our pipeline, yet we did not perform experiments showing if the pipeline is able to provide solutions in terms of the pareto front instead of best performance on a single metric.

	\section{CONCLUSIONS}
	\label{sec:conclusion}

We introduce a gradient-free multi-domain optimization method for autonomous systems in general. The goal of this optimization method is to optimize all system parameters (hardware, software) at the same time to find the desired objective value. We detail the suitable autonomous systems that could be optimized, how to set up the  optimization pipeline, and provide guidance on selecting different variants of the core components in the pipeline according to a specific autonomous system. We also validate our solution by performing a case study on an autonomous race car to show improvements in terms of lap time and compare it to conventional optimization techniques in the same field. By carrying out different experiments we provide insights in the optimization algorithms, hyperparameter tuning, and sensitivity of found solutions. Finally we showed that the proposed method outperforms other methods in the same field for the given task and system.

Future work should focus on performing more case studies using our method on additional autonomous system (e.g. UUV, UAV) with different characteristics. Experiments with combinatorial and discrete parameters should also be performed to further validate the method's ability to handle a wide variety of parameters. Experiments should also consider multi-objective optimization where no objective can be improved without sacrificing at least one other objective. Lastly, future improvements to the method should also consider the reality gap between the simulation evaluations and physical systems.

	\section{Acknowledgment}
	We thank Matthew O'Kelly and Achin Jain during the initial conception of the problem, and Brandon McBride for assistance in development in the software stack.
	
	\appendix
	\begin{figure}[h!]
	\centering
	\includegraphics[scale=0.15]{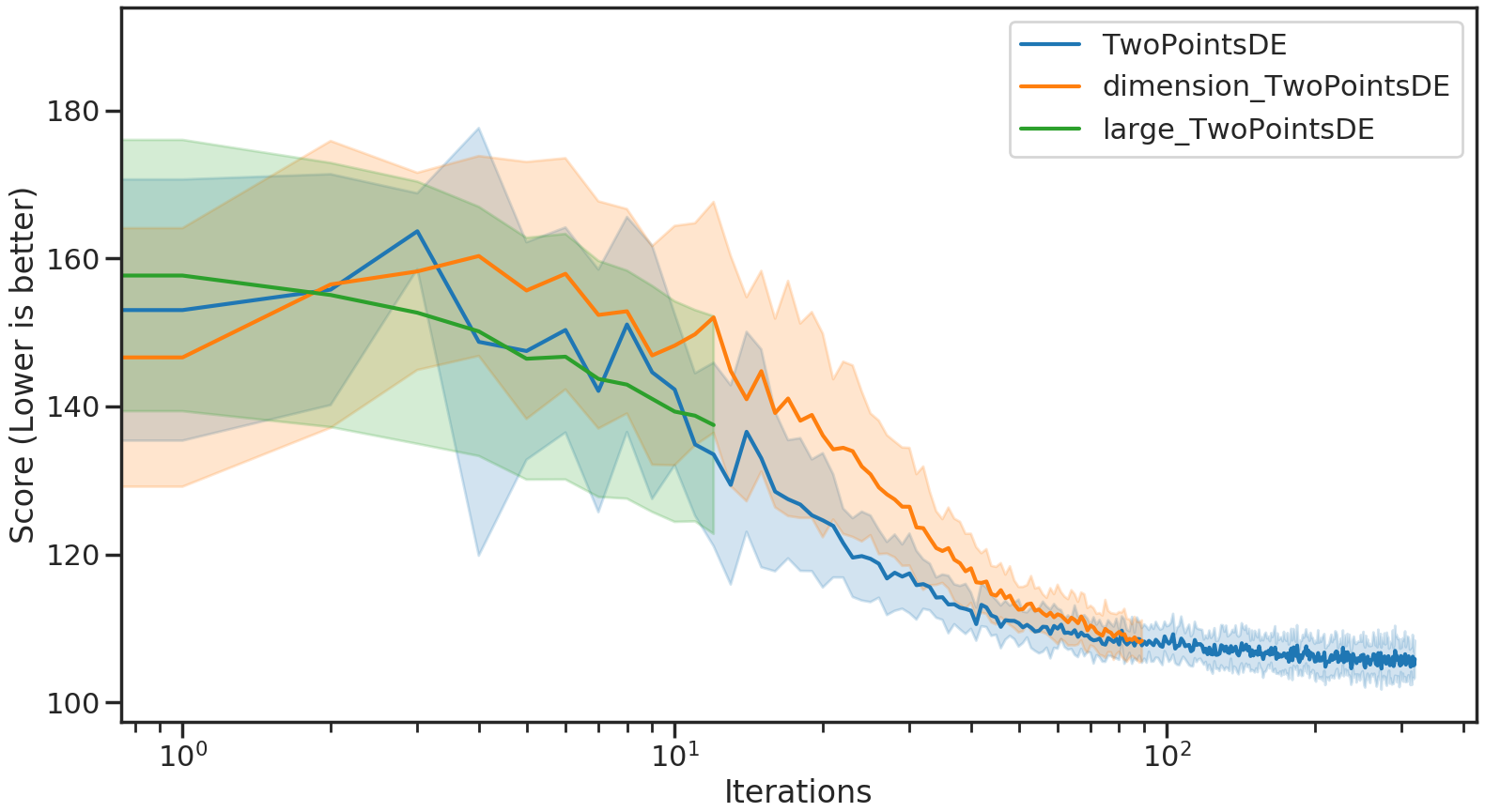}
	\caption{Progression of lap time with different population size used for TwoPointsDE}
	\label{fig:exp3_popsize_2pointsDE}
\end{figure}

\begin{figure}[h!]
	\centering
	\includegraphics[scale=0.15]{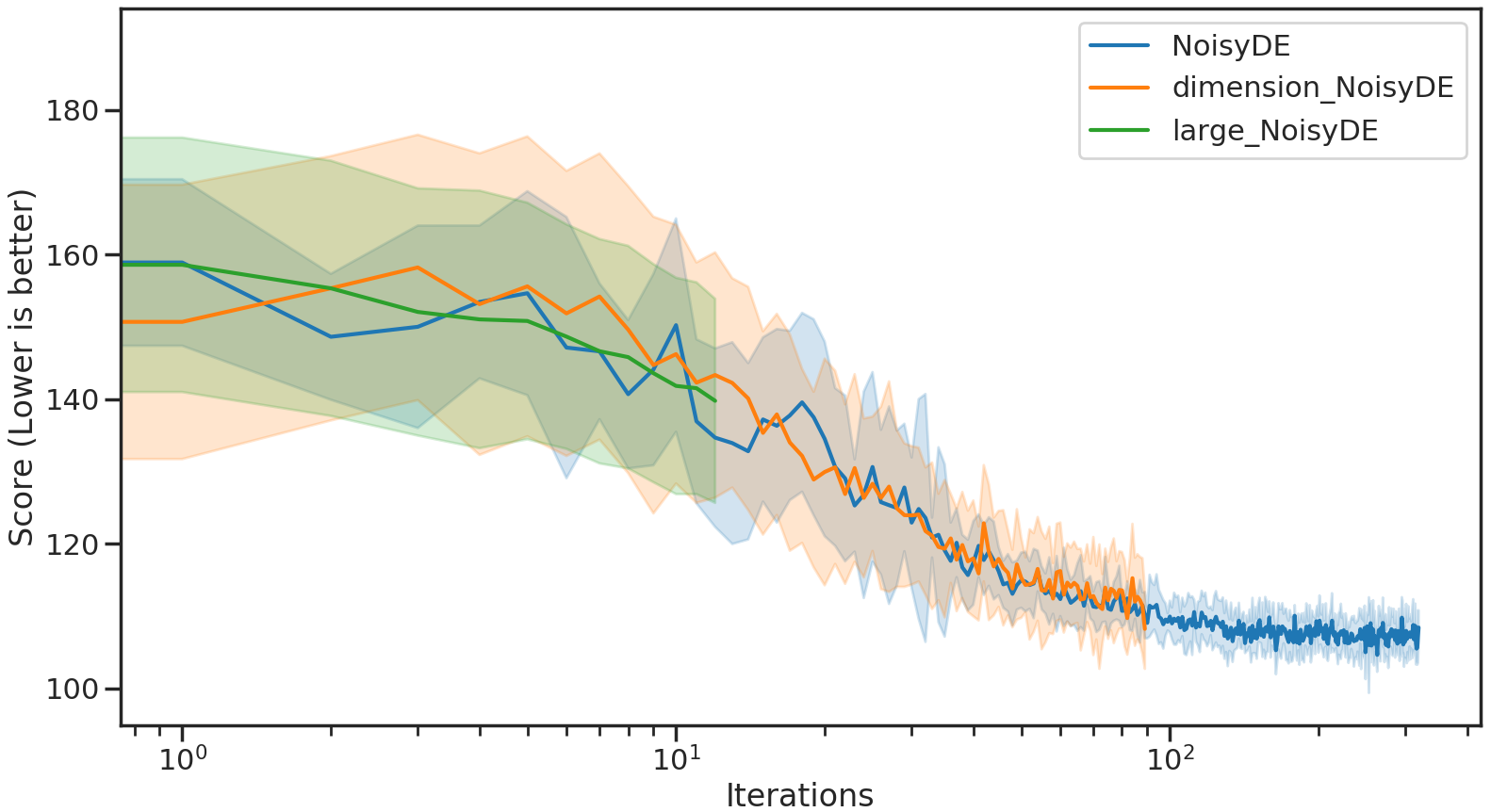}
	\caption{Progression of lap time with different population size used for NoisyDE}
	\label{fig:exp4_popsize_NoisyDE}
\end{figure}

\begin{figure}[h!]
	\centering
	\includegraphics[scale=0.15]{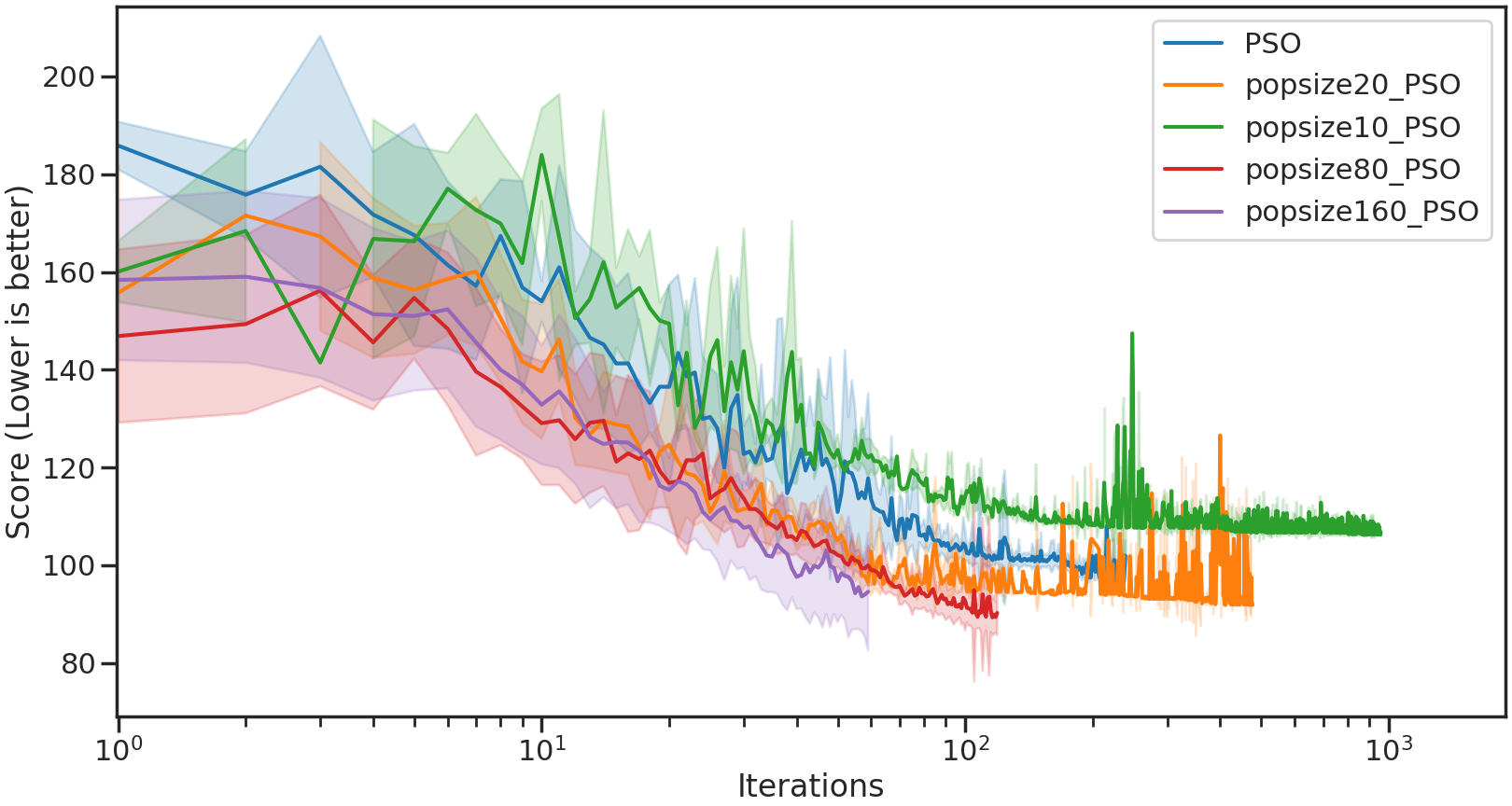}
	\caption{Progression of lap time with different population size used for PSO}
	\label{fig:exp5_popsize_PSO}
\end{figure}

\begin{figure}[h]
	\centering
	\includegraphics[scale=0.08]{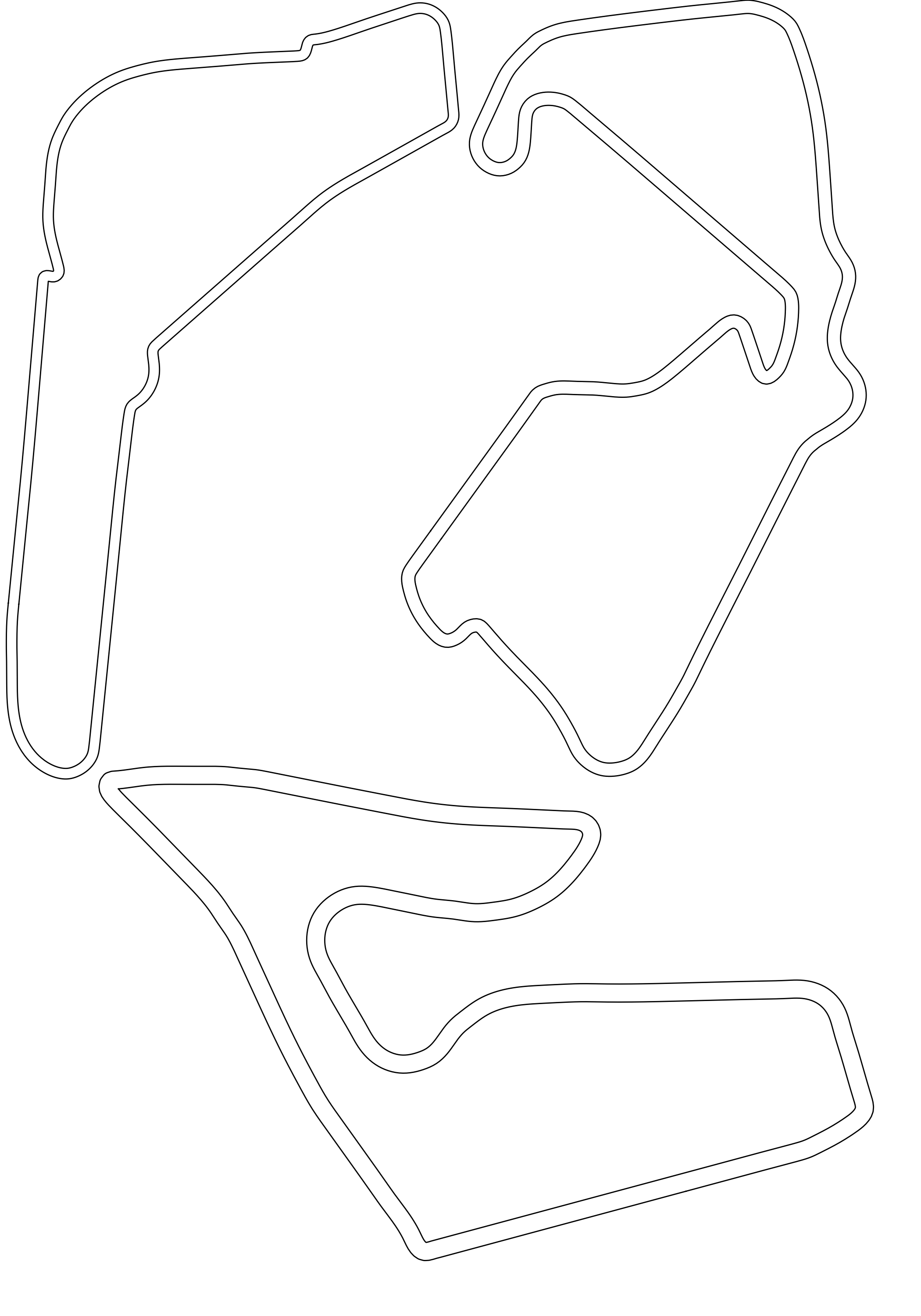}
	\caption{Race track used as environments for experiments. Top Left: Monza, Top Right: Silverstone, Bottom: Spielberg.}
	\label{fig:racetracks}
\end{figure}

\newpage
\begin{table}[h]
	\centering
	\begin{tabular}{|c|c|c|c|}
		\hline
		\textbf{Parameter} & \textbf{\begin{tabular}[c]{@{}c@{}}Pure Pursuit\\ Controller\end{tabular}} & \textbf{\begin{tabular}[c]{@{}c@{}}Stanley\\ Controller\end{tabular}} & \textbf{LQR} \\ \hline
		\textbf{Vehicle Mass (kg)} & 3.9418 & 3.9396 & 3.8052 \\ \hline
		\textbf{CoG to Front (m)} & 0.1610 & 0.1666 & 0.1562 \\ \hline
		\textbf{Max. Velocity (m/s)} & 14.9931 & 11.3578 & 14.9926 \\ \hline
		\textbf{Min. Velocity (m/s)} & 1.1033 & 1.8482 & 1.9779 \\ \hline
		\textbf{Lookahead Distance (m)} & 0.8631 & - & - \\ \hline
		\textbf{Controler Gain kp} & - & 0.0097 & - \\ \hline
		\textbf{Matrix\_Q(1)} & - & - & 0.9991 \\ \hline
		\textbf{Matrix\_Q(2)} & - & - & 0.9537 \\ \hline
		\textbf{Matrix\_Q(3)} & - & - & 0.0066 \\ \hline
		\textbf{Matrix\_Q(4)} & - & - & 0.0257 \\ \hline
		\textbf{Matrix\_R} & - & - & 0.0062 \\ \hline
		\textbf{Lap time (s)} & 66.69 & 92.13 & 62.18 \\ \hline
	\end{tabular}
	\caption{Best path tracker parameters found by CMA on Spielberg.}
	\label{tab:tracker_results}
\end{table}

\begin{sidewaystable}
\tiny
\centering

\begin{tabular}{|c|c|c|c|c|c|c|c|c|c|c|c|c|c|c|c|c|c|c|c|c|c|c|c|c|}
\hline
\multirow{2}{*}{\textbf{Parameters}} & \multicolumn{6}{c|}{\textbf{Comparison of Optimizers}} & \multicolumn{5}{c|}{\textbf{Diff. Pop. size of CMA}} & \multicolumn{3}{c|}{\textbf{Diff. Pop. Size of 2PtsDE \footnote{Both 2PtsDE and NDE has three dimensional settings: standard=max(num\_worker, 30), dimension=max(num\_worker, 30, dimension+1), large=max(num\_worker, 30, 7*dimension)}}} & \multicolumn{3}{c|}{\textbf{Diff. Pop. Size of NDE}} & \multicolumn{5}{c|}{\textbf{Diff. Pop. Size of PSO}} & \multicolumn{2}{c|}{\textbf{Diff. Maps}} \\ \cline{2-25} 
 & \textbf{CMA} & \textbf{2PtsDE} & \textbf{NDE} & \textbf{PSO} & \textbf{1+1} & \textbf{\begin{tabular}[c]{@{}c@{}}Random\\ Search\end{tabular}} & \textbf{6} & \textbf{12} & \textbf{24} & \textbf{48} & \textbf{96} & \textbf{30} & \textbf{106} & \textbf{735} & \textbf{30} & \textbf{106} & \textbf{735} & \textbf{10} & \textbf{20} & \textbf{40} & \textbf{80} & \textbf{160} & \textbf{Silverstone} & \textbf{Monza} \\\hline
\textbf{Vehicle mass (kg)} & 3.94 & 3.47 & 3.93 & 3.76 & 3.57 & 3.87 & 3.29 & 3.32 & 3.94 & 3.82 & 3.90 & 3.47 & 3.59 & 3.78 & 3.93 & 3.83 & 3.91 & 3.51 & 3.98 & 3.76 & 3.76 & 3.56 & 3.96 & 3.94 \\\hline
\textbf{CoG to Front (m) \footnote{Distance from CoG to front axle.}} & 0.16 & 0.15 & 0.17 & 0.15 & 0.17 & 0.16 & 0.15 & 0.16 & 0.16 & 0.15 & 0.15 & 0.15 & 0.16 & 0.16 & 0.17 & 0.16 & 0.16 & 0.16 & 0.16 & 0.15 & 0.17 & 0.16 & 0.15 & 0.15 \\\hline 
\textbf{Max. Velocity (m/s)} & 14.99 & 9.78 & 9.65 & 10.62 & 10.83 & 9.32 & 15.00 & 13.92 & 14.99 & 14.89 & 14.10 & 9.78 & 9.93 & 9.79 & 9.65 & 10.73 & 10.35 & 10.22 & 10.63 & 10.62 & 11.24 & 11.88 & 14.82 & 10.42 \\\hline 
\textbf{Min. Velocity (m/s)} & 1.10 & 1.47 & 1.99 & 1.47 & 0.60 & 1.93 & 1.37 & 1.22 & 1.10 & 1.55 & 1.73 & 1.47 & 1.31 & 1.92 & 1.99 & 0.96 & 1.90 & 1.21 & 1.99 & 1.47 & 1.94 & 1.59 & 1.90 & 1.67 \\\hline 
\textbf{\begin{tabular}[c]{@{}c@{}}Lookahead \\ Distance (m)\end{tabular}} & 1.75 & 1.00 & 1.00 & 1.00 & 1.17 & 2.00 & 1.66 & 1.43 & 1.75 & 1.86 & 1.82 & 1.00 & 1.19 & 1.19 & 1.00 & 1.87 & 1.71 & 0.94 & 1.17 & 1.01 & 1.12 & 1.46 & 1.46 & 1.09 \\\hline 
\textbf{Laptime (s)} & 66.69 & 100.20 & 99.85 & 96.39 & 110.21 & 115.29 & 69.65 & 73.71 & 66.69 & 67.90 & 73.78 & 100.20 & 101.09 & 108.63 & 99.85 & 98.52 & 106.47 & 106.20 & 91.86 & 96.39 & 86.97 & 85.91 & 89.56 & 113.04 \\\hline 
\end{tabular}
\caption{Overview of optimized parameters from different experiments using Pure Pursuit}
\label{tab:all_param}
\end{sidewaystable}
	
	\clearpage
	\bibliographystyle{IEEEtran}
	\bibliography{billy.bib,brandon.bib}
	
	\begin{IEEEbiography}[{\includegraphics[width=1in,height=1.25in,clip,keepaspectratio]{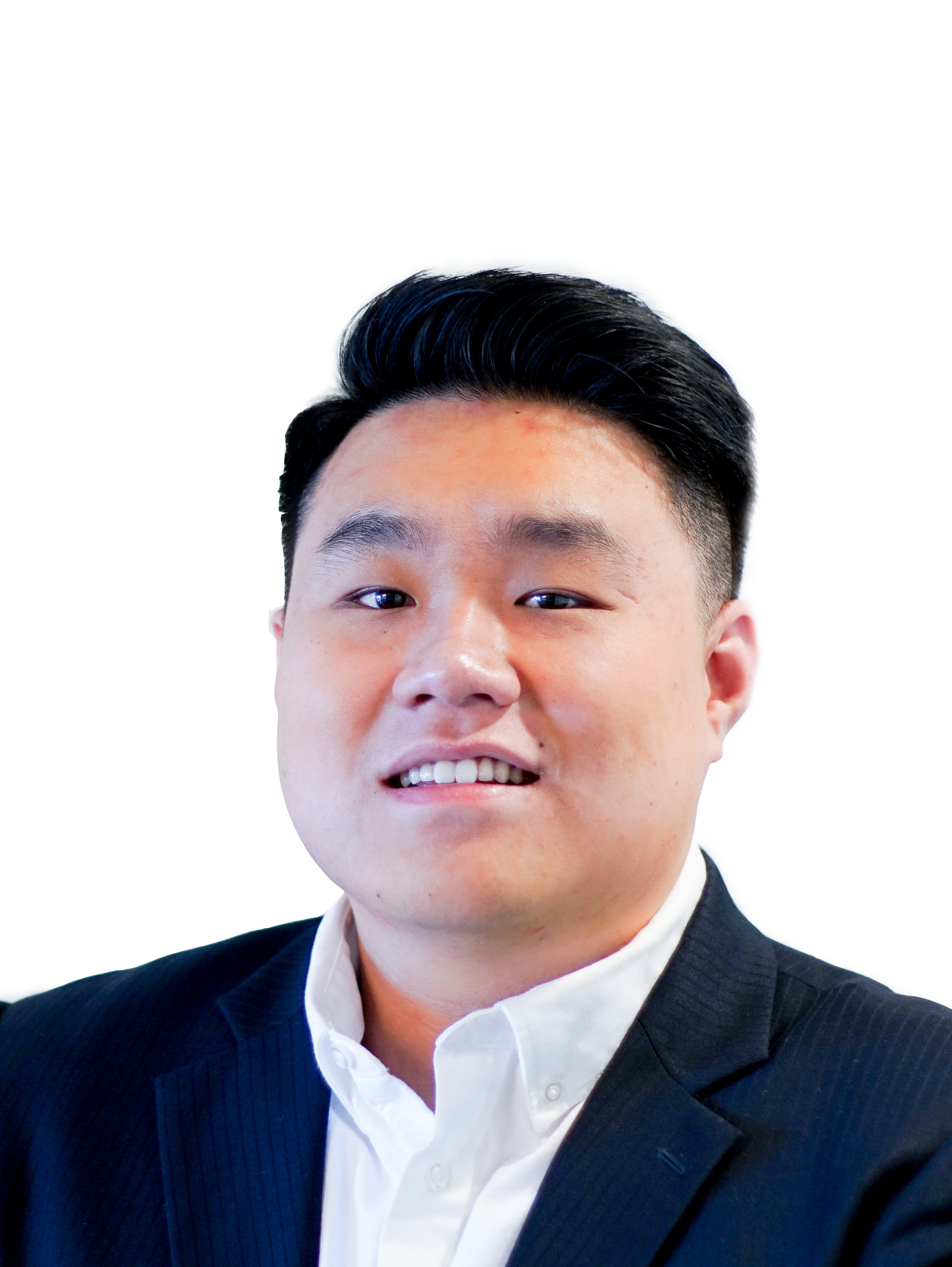}}]{Hongrui Zheng} received his B.S. degrees in Mechanical Engineering and Computer Science from Georgia Institute of Technology, and his M.S. degree in Robotics from University of Pennsylvania. He is currently a Ph.D. candidate at University of Pennsylvania. His research focuses on building the tools and theoretical foundations necessary to scale design, testing, and optimization of safe-autonomous systems.
\end{IEEEbiography}

\begin{IEEEbiography}[{\includegraphics[width=1in,height=1.25in,clip,keepaspectratio]{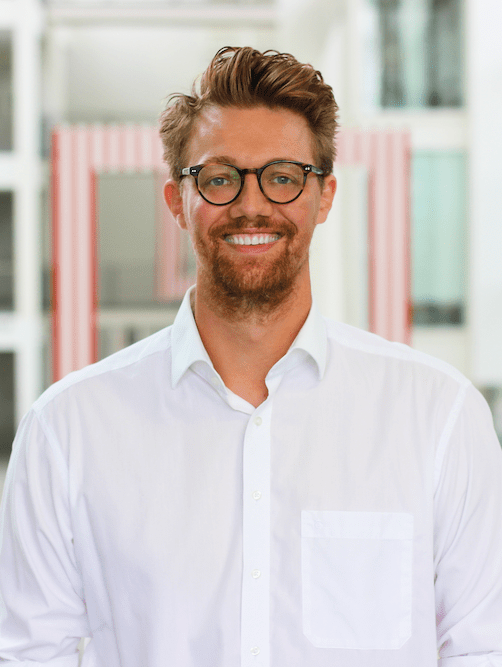}}]{Johannes Betz} is a postdoctoral researcher at the University of Pennsylvania where he is am working at the mLab: Real-Time and Embedded Systems Lab. His research is focusing on a holistic software development for autonomous systems with extreme motions at the dynamic limits in extreme and unknown environments. By using modern algorithms from the field of artificial intelligence he is trying to develop new and advanced methods and intelligent algorithms. Based on his additional studies in philosophy he extends current path and behavior planners for autonomous systems with ethical theories.
\end{IEEEbiography}

\begin{IEEEbiography}[{\includegraphics[width=1in,height=1.25in,clip,keepaspectratio]{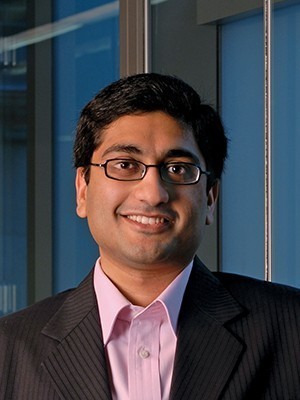}}]{Rahul Mangharam} received his Ph.D. in Electrical and Computer Engineering from Carnegie Mellon University where he also received his MS and BS.  He is an Associate Professor in the Department of Electrical and Systems Engineering at the University of Pennsylvania. He is a founding member of the PRECISE Center and directs the Safe Autonomous Systems Lab at Penn. His research is at the intersection of formal methods,  machine learning and control for medical devices,  energy efficient buildings, and autonomous systems.
\end{IEEEbiography}

\end{document}